\documentclass[a4paper,UKenglish,cleveref, autoref]{lipics-v2019}
\usepackage{booktabs}
\usepackage{url}
\usepackage{float}
\usepackage[compact]{titlesec}

\title{LSTM-TrajGAN: A Deep Learning Approach to Trajectory Privacy Protection} 

\titlerunning{LSTM-TrajGAN}

\author{Jinmeng Rao}{Geospatial Data Science Lab, University of Wisconsin-Madison, USA}{}{https://orcid.org/0000-0003-2370-5129}{}

\author{Song Gao}{Geospatial Data Science Lab,  University of Wisconsin-Madison, USA}{}{http://orcid.org/0000-0003-4359-6302}{}

\author{Yuhao Kang}{Geospatial Data Science Lab, University of Wisconsin-Madison, USA}{}{https://orcid.org/0000-0003-3810-9450}{}

\author{Qunying Huang}{Department of Geography, University of Wisconsin-Madison, USA}{}{https://orcid.org/0000-0003-3499-7294}{}

\authorrunning{Rao et al. 2020}

\Copyright{John Q. Public and Joan R. Public}

\ccsdesc[500]{Security and privacy~Privacy protections}
\ccsdesc[500]{Computing methodologies~Artificial intelligence}

\keywords{GeoAI, Deep Learning, Trajectory Privacy, Generative Adversarial Networks}

\category{}

\relatedversion{}

\supplement{}

\acknowledgements{Support for this research was provided by the University of Wisconsin - Madison Office of the Vice Chancellor for Research and Graduate Education with funding from the Wisconsin Alumni Research Foundation.}

\nolinenumbers 

\hideLIPIcs  

\EventEditors{Krzysztof Janowicz and Judith Verstegen}
\EventNoEds{2}
\EventLongTitle{11th International Conference on Geographic Information Science (GIScience 2020)}
\EventShortTitle{GIScience 2020}
\EventAcronym{GIScience}
\EventYear{2020}
\EventDate{September 15--18, 2020}
\EventLocation{Poznań, Poland}
\EventLogo{}
\SeriesVolume{42}
\ArticleNo{1}
\begin{document}

\maketitle

\begin{abstract}
\footnote{a preprint and the final version will be available in the Proceedings of the 11th International Conference on Geographic Information Science (GIScience 2021) \url{https://www.giscience.org/}.} The prevalence of location-based services contributes to the explosive growth of individual-level trajectory data and raises public concerns about privacy issues. In this research, we propose a novel LSTM-TrajGAN approach, which is an end-to-end deep learning model to generate privacy-preserving synthetic trajectory data for data sharing and publication. We design a loss metric function TrajLoss to measure the trajectory similarity losses for model training and optimization. The model is evaluated on the trajectory-user-linking task on a real-world semantic trajectory dataset. Compared with other common geomasking methods, our model can better prevent users from being re-identified, and it also preserves essential spatial, temporal, and thematic characteristics of the real trajectory data. The model better balances the effectiveness of trajectory privacy protection and the utility for spatial and temporal analyses, which offers new insights into the GeoAI-powered privacy protection.
\end{abstract}

\section{  Introduction} \label{sec:intro}
The increasing location-based services (LBS) have generated large-scale individual-level trajectory data (i.e., a sequence of locations with attributes) through mobile phones, wearable sensors, GPS devices, and geotagged social media \cite{liu2015social}. Such trajectory big data provide new opportunities to study human mobility patterns and human-environment interactions \cite{huang2015modeling}, disaster responses \cite{huang2015geographic,wang2016spatial} and public health issues \cite{li2019dynamic,park2017individual}. It also introduces grand challenges regarding the protection of geoprivacy and broader behavioral, social, ethical, legal and policy implications \cite{kessler2018geoprivacy}. Generally speaking, trajectory privacy refers to an individual's rights to prevent the disclosure of individual trajectory identity and associated personal sensitive locations \cite{kwan2004protection,chow2011trajectory,gao2019exploring}.

Due to the data breach concerns and increasing public awareness of location privacy protection, many approaches have been proposed to prevent users' trajectories from being identified. A common practice is to remove the identifiers (e.g., user name or ID number) from the trajectory data. However, it turned out that such "de-identified" trajectories may still cause serious privacy threats since the spatial,  temporal and thematic characteristics of trajectories can still be used as strong quasi-identifiers for linking the trajectories to their creators \cite{chow2011trajectory}. Another commonly used method is to aggregate trajectory points into geographic or administrative units so that their original locations are not revealed. Nevertheless, recent studies show that aggregation may not only fail to preserve user privacy, but also reduce the spatial resolution and effectiveness of spatial analysis \cite{de2013unique, xu2017trajectory, gao2019exploring}. For example, De Montjoye et al. \cite{de2013unique} lower the resolution of a human mobility trace dataset through spatial and temporal aggregation to preserve the individuals from being identified, but the coarsened dataset still provides little anonymity. Thus, in order to achieve trajectory privacy protection more efficiently, we need to deal with the spatial and temporal characteristics of trajectory data more specifically.

Current trajectory privacy protection studies focus on two research streams. One is the differential privacy approach to grouping and mixing the trajectories from different users so that the identification of individual trajectory data is converted into a k-anonymity problem \cite{niu2014achieving,zhu2019private}. For example, the spatial cloaking approach mixes together the trajectory points between k users using k-anonymous cloaked spatial regions, making these trajectories k-anonymized \cite{gruteser2003anonymous}. Also, the mix-zones approach anonymizes the trajectory points in a mix-zone using pseudonyms and breaks the linkage between the former segment and the latter segment of the same trajectory that passes through a mix-zone \cite{palanisamy2011mobimix}. Alternatively, the generalization-based approach first divides the points of $k$ trajectories into different k-anonymized regions, and then reconstructs $k$ new trajectories by uniformly selecting points from each k-anonymized regions and linking them together \cite{nergiz2008towards}. 

Another research stream is called geomasking, which blurs the locations of original trajectory data by utilizing perturbation on the spatial dimension so that the original locations can be hidden or modified while spatial patterns may not be significantly affected \cite{hampton2010mapping, gao2019exploring}. For example, Armstrong et al. \cite{armstrong1999geographically} explored the privacy preservation ability and spatial analysis effectiveness of several types of geomasks. Kwan et al. \cite{kwan2004protection} evaluated the spatial analysis effectiveness of three different random perturbation geomasks on lung-cancer deaths. Seidl et al. \cite{seidl2016privacy} applied grid masking and random perturbation on GPS trajectory data and evaluated the privacy protection performance. Gao et al. \cite{gao2019exploring} investigated the effectiveness of random perturbation, gaussian perturbation, and aggregation on Twitter data as well as explored the privacy, analytics, and uncertainty level of each method.

While these approaches all show the capabilities to protect trajectory privacy, they also expose several limitations. First of all, despite the diversity, the goal of these approaches largely is to obfuscate the trajectory locations and add more uncertainty to preserve privacy. However, the trade-off between the effectiveness of trajectory privacy protection and the utility for spatial and temporal analyses is still hard to control \cite{liu2018trajgans}, and this issue has not been fully discussed or evaluated. Besides, current studies mainly focus on the spatial dimension of trajectory data whereas other semantics (e.g., temporal and thematic attributes) are rarely considered. In fact, these characteristics have been proven to be crucial for trajectory user identification \cite{may2020marc}. Moreover, current approaches rely heavily on manually designed procedures. Once the procedure is disclosed, one may have the chance to recover the original trajectory data \cite{xu2017trajectory} (e.g., using reverse engineering). The "black-box" machine learning models may help to solve this issue.

To this end, this research aims to explore the effectiveness of state-of-the-art deep learning approaches for trajectory privacy protection. We propose a novel LSTM-TrajGAN model that combines the Long Short-Term Memory (LSTM) recurrent neural network and the Generative Adversarial Network (GAN) structure together to generate privacy-preserving synthetic trajectories as alternatives to real trajectories for trajectory data sharing and publication. Two research questions (RQ) will be investigated in this work. 

\par {RQ 1:} How effective is the proposed LSTM-TrajGAN model in protecting the trajectory creators from being re-identified? (i.e., privacy protection effectiveness)
\par {RQ 2:} Can the synthetic trajectories preserve the semantic features (spatial-temporal-thematic characteristics) compared to real trajectories? (i.e., utility)

The main contributions of our work are fourfold: (1) we propose an end-to-end deep learning approach to generating privacy-preserving trajectory data. The procedure is simple and highly secure (a GeoAI "black-box"); (2) we introduce a trajectory encoding model for semantic trajectory encoding; (3) we design a new TrajLoss metric function to measure the trajectory similarity losses for training deep learning models; and (4) we evaluate the privacy protection effectiveness and the utility of the proposed model using real-world LBS data and explore the trade-off between them.

The remainder of the paper is organized as follows. Section 2 introduces our methodological framework, including a trajectory encoding model, the LSTM-TrajGAN model, and the TrajLoss function design. In section 3, we train and test our model using a city-scale weekly trajectory dataset and compare with other commonly used trajectory privacy protection methods. Both privacy protection effectiveness and utility are evaluated and compared with baseline approaches; In section 4, we discuss the factors affecting privacy protection effectiveness, the trade-off between privacy protection and utility, and the limitations of our model. Section 5 summarizes this research and outlines the future work.
\section{  Method} \label{sec:method}
Inspired by the vision of the TrajGANs \cite{liu2018trajgans}, we propose a new approach consisting of three main components: (1) a Trajectory Encoding Model, which encodes GPS location coordinates, temporal attributes, and other attributes such as point of interest (POI) category; (2) a Trajectory Generator, which takes random noise and original trajectories as inputs to generate synthetic trajectories as outputs; and (3) a Trajectory Discriminator, which takes trajectories as inputs and determines them as "real" or "synthetic". 

\begin{figure}[h]
  \centering
  \includegraphics[width=0.9\linewidth]{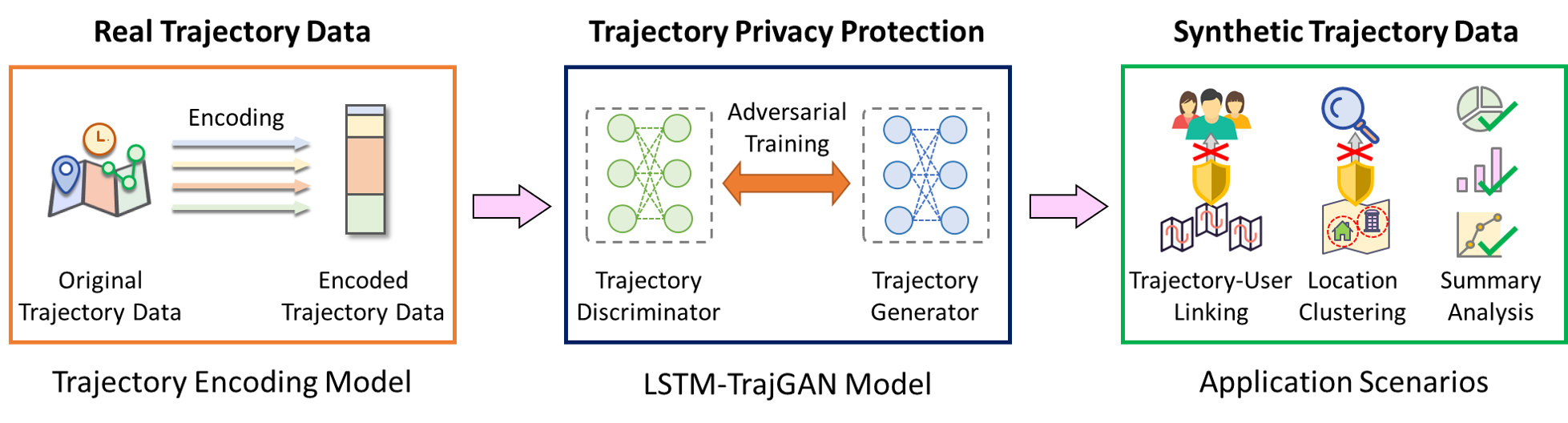}
  \caption{The overall workflow of the proposed LSTM-TrajGAN approach.}
  \label{fig:workflow}
\end{figure}

The overall workflow is described in Figure \ref{fig:workflow}. The goal is to train an "intelligent" trajectory generator that generates "realistic" synthetic trajectories to replace the original trajectories, which preserves differential privacy in trajectory analysis tasks such as Trajectory-User Linking (TUL) and trajectory data mining (e.g., work/home location clustering). Meanwhile, it ensures the quality of multiple spatial or temporal summary analysis tasks. Such a framework can serve as a trajectory privacy protection layer in trajectory data acquisition, processing, and publication pipelines, which publish the synthetic alternatives rather than the real trajectory data that may disclose individual privacy.

\subsection{  Trajectory Encoding}
First, we introduce a trajectory encoding model that converts the original trajectory data to a specific format that serves as the inputs for the LSTM-TrajGAN model. The main reason for the encoding process is that the trajectory data usually contain various types of attributes such as interval data (e.g., GPS coordinates, date and time), nominal data (e.g., POI category), ordinal data (e.g., POI rating), and these data need to be converted into valid numerical representations for training the deep learning model. Our trajectory encoding model includes two parts: trajectory point encoding and trajectory padding.

\begin{figure}[h]
  \centering
  \includegraphics[width=0.8\linewidth]{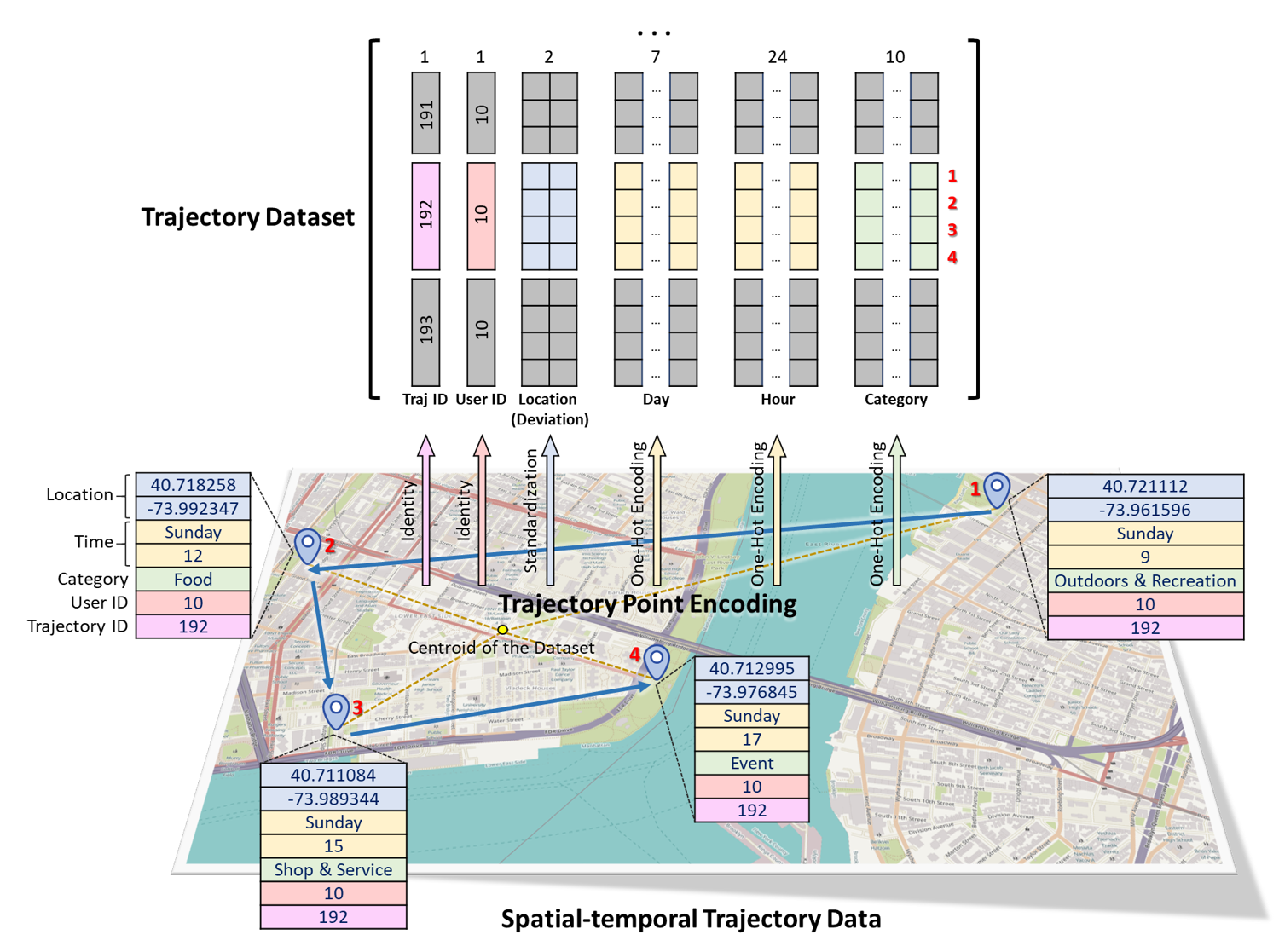}
  \caption{An example for the trajectory point encoding process.}
  \label{fig:trajectory_point_encoding}
\end{figure}

\paragraph*{Trajectory Point Encoding}
The trajectory point encoding process is illustrated in Figure \ref{fig:trajectory_point_encoding}. A semantic trajectory point contains the following attributes: location, time, user id, trajectory id, and other optional attributes such as POI category. For the location attribute, we standardize all the latitudes and longitudes using the centroid of all the trajectories in the dataset to obtain the deviations of the latitudes and longitudes from the centroid. In this way, the model can better learn the spatial deviation pattern between different trajectory points. These deviation values will be used as the numerical representations of the trajectory points for constructing spatial embeddings \cite{space2vec_iclr2020}.

For the temporal attributes and categorical attributes, we use the one-hot encoders (i.e., a representation process using dummy variables in machine learning) to encode the attributes into high-dimensional binary vectors based on their vocabulary sizes. For example, the "Day" attribute is encoded into 7-dimensional binary vectors, and "Monday" is represented as $[1, 0, 0, 0, 0, 0, 0]$. Likewise, the "Hour" attribute is encoded into 24-dimensional binary vectors, and the "Category" attribute is encoded into 10-dimensional binary vectors. Note that we don't encode the User ID and the Trajectory ID since they are only used to indicate the user and the trajectory that the point belongs to.

\paragraph*{Trajectory Padding}

After the trajectory point encoding process, all the spatial, temporal, and thematic attributes of a trajectory are stored in a multidimensional matrix, whose first dimension indicates the index for each trajectory. Since the length of each trajectory data (i.e., the number of the trajectory points) is a variable, we then apply the trajectory padding technique to make sure all the trajectories have the same length as the longest trajectory. Specifically, we use zero pre-padding to pad empty trajectory points (i.e., the points whose attributes are all set to zero) to each trajectory until all the trajectories reached the same length as the longest trajectory in the dataset. The main reason is that the data with the same size can be utilized for batch processing and training the deep learning model, which would speed up the training process. During the model training and inference processes, these padded trajectory points will be masked (i.e., cut) and they won't actually influence the neural network weight updates and the derived results.

\subsection{  LSTM-TrajGAN Model}
\begin{figure}[h]
  \centering
  \includegraphics[width=0.8\linewidth]{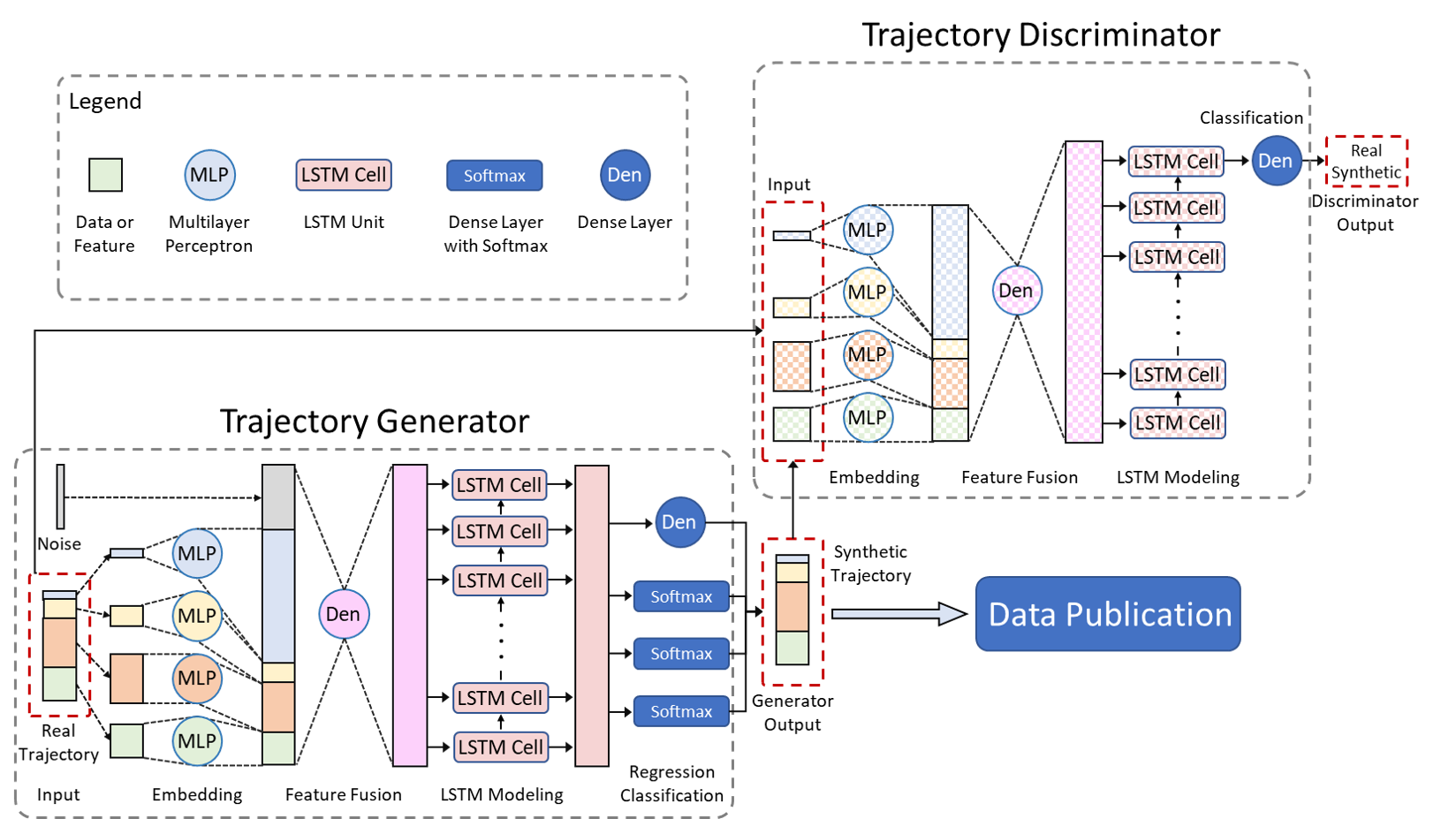}
  \caption{The neural network structure of the LSTM-TrajGAN Model.}
  \label{fig:lstm_trajgan}
\end{figure}

Figure \ref{fig:lstm_trajgan} describes the neural network structure of the LSTM-TrajGAN Model. The trajectory generator captures the data distribution and pattern of the real trajectory data and generates synthetic trajectory data based on their corresponding original trajectory data and random noise. In addition, the trajectory discriminator distinguishes whether the trajectory samples come from the training set (i.e., real trajectory data) or the trajectory generator (i.e., synthetic trajectory data). The goal of the trajectory generator is to generate "high-quality" synthetic trajectories that can "fool" the trajectory discriminator, which leads to a two-player minimax game between them. The generated synthetic trajectories aim to be competent for spatial and temporal summary analysis,  while having some degree of uncertainty and randomness to protect the user privacy in trajectory analysis tasks with privacy issues involved. This idea is reflected in the design and optimization of the LSTM-TrajGAN model.

\paragraph*{Trajectory Generator}
As is shown in Figure \ref{fig:lstm_trajgan}, the trajectory generator consists of five functional layers: the input layer, the embedding layer, the feature fusion layer, the LSTM modeling layer, and the regression/classification layer. The generator first takes the encoded real trajectories and random noise as inputs, and embeds trajectories using Multilayer Perceptrons (MLPs) \cite{haykin1994neural}. For the spatial dimension of a trajectory (i.e., pairs of latitude and longitude deviations), we embed each pair of them using a MLP to get 64-dimensional vectors. For the temporal dimension (e.g., day and hour) and categorical attributes (e.g., POI category), we use MLPs to embed them respectively and get fixed-length vectors based on their vocabulary sizes:

\begin{equation}
e^{spatial}_{i} = \phi^{s}(\Delta lat_i,\Delta lon_i; W_{es})
\end{equation}
\begin{equation}
e^{day}_{i} = \phi^{d}(v^{day}_{i}; W_{ed})
\end{equation}
\begin{equation}
e^{hour}_{i} = \phi^{h}(v^{hour}_{i}; W_{eh})
\end{equation}
\begin{equation}
e^{category}_{i} = \phi^{c}(v^{category}_{i}; W_{ec})
\end{equation}

Where $\Delta lat_i$ and $\Delta lon_i$ stand for the latitude and longitude deviation of the i-th trajectory point; $v^{day}_{i}$, $v^{hour}_{i}$, $v^{category}_{i}$ stand for the one-hot vectors for the day, hour, and category attributes of the i-th trajectory point; $\phi^{s}$, $\phi^{d}$, $\phi^{h}$, and $\phi^{c}$ stand for the MLPs with an activation function -- the Rectified Linear Unit (ReLU) for embedding the spatial, daily, hourly, and categorical attributes; $W_{es}$, $W_{ed}$, $W_{eh}$, and $W_{ec}$ are the embedding weight matrices for these MLPs; $e^{spatial}_{i}$, $e^{day}_{i}$, $e^{hour}_{i}$, and $e^{category}_{i}$ are the embedded vectors for each attribute respectively. Note that the embedding weight matrices are shared among all trajectory points.

After the embedding process, we further concatenate all the vectors and the random noise, and then use a dense layer to fuse them into 100-dimensional vectors. By leveraging the feature fusion, we take the advantage of all the spatial, temporal, and categorical characteristics of each trajectory point and fuse them together to support spatiotemporal trajectory modeling and generation.
In the LSTM modeling layer, we use a many-to-many LSTM structure that takes a sequence with specific time steps as the input and generates a sequence with the same time steps as the output. Recurrent models such as LSTM are proven to be efficient in spatial-temporal sequence modeling and prediction \cite{gupta2018social,may2020marc}. Given the dimension of the fused feature, we assign 100 units in the LSTM model and feed the fused features to the model:

\begin{equation}
H = LSTM(F; W_{lstm})
\end{equation}

Where $F$ represents for the fused features of all the trajectory points in a trajectory (i.e., $ F = [f_{0}, f_{1}, ..., f_{maxlength - 1}]$), in which $f_{i}$ is the fused feature vector for the i-th trajectory point; $H$ is the output of the LSTM model, which has the same time step dimensions as the input (i.e., $ H = [h_{0}, h_{1}, ..., h_{maxlength - 1}]$, in which $h_{i}$ is the modeling output vector for $f_{i}$); $W_{lstm}$ is the weight matrix of the LSTM model.

Finally, we decode the synthetic trajectory data from the output $H$ of the LSTM modeling layer. Each feature vector $h_{i}$ in $H$ is a 100-dimensional vector containing the spatial, temporal, and categorical characteristics of a synthetic trajectory point. To decode the latitude and longitude deviations, we use a dense layer with two units and use the $tanh$ hyperbolic tangent function. In addition, we further stretch the output range to make sure its range covers all the possible deviation values. To decode the day, hour and category attributes, we use dense layers that have as many units as the vocabulary sizes, and use the $softmax$ normalized exponential function to recover the one-hot representation of these attributes:

\begin{equation}
(\Delta lat'_i,\Delta lon'_i) = D^{s}(h_{i}; W_{ds})
\end{equation}
\begin{equation}
v'^{day}_{i} = D^{d}(h_{i}; W_{dd})
\end{equation}
\begin{equation}
v'^{hour}_{i} = D^{h}(h_{i}; W_{dh})
\end{equation}
\begin{equation}
v'^{category}_{i} = D^{c}(h_{i}; W_{dc})
\end{equation}

Where $\Delta lat'_i$ and $\Delta lon'_i$ are the latitude and longitude deviations of the i-th synthetic trajectory point; $v'^{day}_{i}$, $v'^{hour}_{i}$, $v'^{category}_{i}$ represent the one-hot vectors for the day, hour, and category attributes of the i-th synthetic trajectory point; $D^{s}$, $D^{d}$, $D^{h}$, and $D^{c}$ represent the dense layers with a $tanh$ or $softmax$ function for decoding the location, day, hour, and category attributes; $W_{ds}$, $W_{dd}$, $W_{dh}$, and $W_{dc}$ are the decoding weight matrices for these dense layers; Note that the decoding weight matrices are shared among all trajectory points.

\paragraph*{Trajectory Discriminator}

As is shown in Figure \ref{fig:lstm_trajgan}, the trajectory discriminator has a very similar structure as the trajectory generator. The major differences between them are:

(1) The discriminator only takes trajectory data as the input (no random noise needed);

(2) We use a many-to-one LSTM model that takes the features with time steps as the input and make one scalar as the output:

\begin{equation}
h = LSTM(F; W_{lstm_d})
\end{equation}

Where F represents for the fused features of all the trajectory points in a trajectory (i.e., $ F = [f_{0}, f_{1}, ..., f_{maxlength - 1}]$), in which $f_{i}$ is the fused feature vector for the i-th trajectory point; $W_{lstm_d}$ is the weight matrix of the LSTM model; and $h$ is the output scalar of the LSTM model.

(3) We use a one-unit dense layer with the $sigmoid$ activation function to make binary classification (real or synthetic) on the scalar output:

\begin{equation}
O_d = D^{bc}(h; W_{bc})
\end{equation}

Where $D^{bc}$ is the one-unit dense layer with a $sigmoid$ function used to make binary classification, and $W_{bc}$ is its weight matrix; $O_d$ is the final output of the discriminator.  

\subsection{  TrajLoss for Measuring Trajectory Similarity Losses}

The original GAN is designed to optimize the following objective function \cite{goodfellow2014generative}:

\begin{equation}
V(D,G)=\min\limits_G \max\limits_D (\mathbb{E}_{x\sim p_{data}(x)}[logD(x)] + \mathbb{E}_{z\sim p_{z}(z)}[log(1-D(G(z)))])
\end{equation}

Where $p_{data}(x)$ represents the distribution of the real data samples; $p_{z}(z)$ represents a prior on noise variables; $D(x)$ represents the probability
that $x$ came from $p_{data}(x)$; $G(z)$ represents a mapping from $p_{z}(z)$ to $p_{data}(x)$. The generator aims to minimize $\mathbb{E}_{z\sim p_{z}(z)}[log(1-D(G(z)))]$ while the discriminator aims to maximize $\mathbb{E}_{x\sim p_{data}(x)}[logD(x)] + \mathbb{E}_{z\sim p_{z}(z)}[log(1-D(G(z)))]$, leading to a two-player minimax game.

According to the objective function $V(D,G)$, the loss function for the discriminator can be considered as a Binary Cross-Entropy (BCE) loss function ($L_{BCE}$), which will also be used in training the generator. However, different from the original GAN, we need the real trajectory data as inputs. Thus, we design a new loss metric function named TrajLoss to further measure the similarity losses between the real trajectory data and the synthetic trajectory data in spatial, temporal and categorical dimensions, and use this loss function to train the generator. The TrajLoss is defined as follows:
\begin{equation}
TrajLoss (y^{r}, y^{p}, t^{r}, t^{s}) = \alpha L_{BCE}(y^{r},y^{p}) + \beta L_{s}(t^{r}, t^{s}) + \gamma L_{t}(t^{r}, t^{s}) + c L_{c}(t^{r}, t^{s})
\end{equation}

Where $y^{r}$ and $y^{p}$ represent the ground truth label and the prediction result of the trajectory by the discriminator, respectively; $t^{r}$ and $t^{s}$ represent the real trajectory and the corresponding synthetic trajectory; $L_{BCE}$ is the original binary cross-entropy loss from the discriminator; $L_{s}$, $L_{t}$, and $L_{c}$ are the spatial similarity loss, temporal similarity loss, and the categorical similarity loss between the real and synthetic trajectories, respectively; $\alpha$, $\beta$, $\gamma$, and $c$ are the weights for these losses and can be assigned differently for different scenarios.

In this paper, we use the L2 loss (i.e., least square errors) for $L_{s}$ as a recent study \cite{gupta2018social} shows that the L2 loss is effective in measuring trajectory spatial similarity. Besides, we choose the Softmax Cross-Entropy (SCE) as the loss function for $L_{t}$ and $L_{c}$ since they are all regarded as multi-classification problems in this framework, and thus can be optimized using SCE. During the model training, the weights of the generator will be updated by the TrajLoss to improve the quality of the synthetic trajectory data.

\section{  Experiments}
To address the abovementioned RQ1, this section first evaluates the effectiveness of trajectory privacy protection using the proposed LSTM-TrajGAN model on a classic LBS task: Trajectory-User Linking (TUL), which identifies users from trajectories and link trajectories to them \cite{gao2017identifying}. TUL is an essential task in geo-tagged social media applications and receives increasing privacy concerns \cite{gao2017identifying,zhou2018trajectory,may2020marc}. The evaluation can be regarded as an adversarial experiment: we train the LSTM-TrajGAN model and use the generated synthetic trajectories to suppress the accuracy of a state-of-the-art TUL algorithm. We also compare our approach with the other two commonly used location privacy protection methods: Random Perturbation and Gaussian Geomasking.

Meanwhile, to address the RQ2 for verifying the utility of the proposed model (i.e., the usefulness of the synthetic trajectories in analysis), we also explore the spatial and temporal characteristics of the synthetic trajectories to see if they preserve sufficient information from the original trajectories to further support spatial and temporal analyses.

\subsection{  Trajectory-User Linking}

\paragraph*{Dataset}

We use the Foursquare weekly trajectory dataset in New York City (NYC) provided by Petry et al. \cite{may2020marc}, which is extracted from the Foursquare NYC check-ins dataset \cite{yang2014modeling}. We only keep the user ID, trajectory ID, location, hour, day, and category attributes and remove other attributes (e.g., price tier, rating, weather). The summary of the attributes is shown in Table \ref{table:tuldataset}. There are 193 users, 3,079 trajectories and 66,962 trajectory points in total in the dataset. We use 2/3 of the trajectories for training the LSTM-TrajGAN model and 1/3 for testing as suggested in \cite{may2020marc}.

\begin{figure}[!htb]
  \centering
  \includegraphics[width=0.8\linewidth]{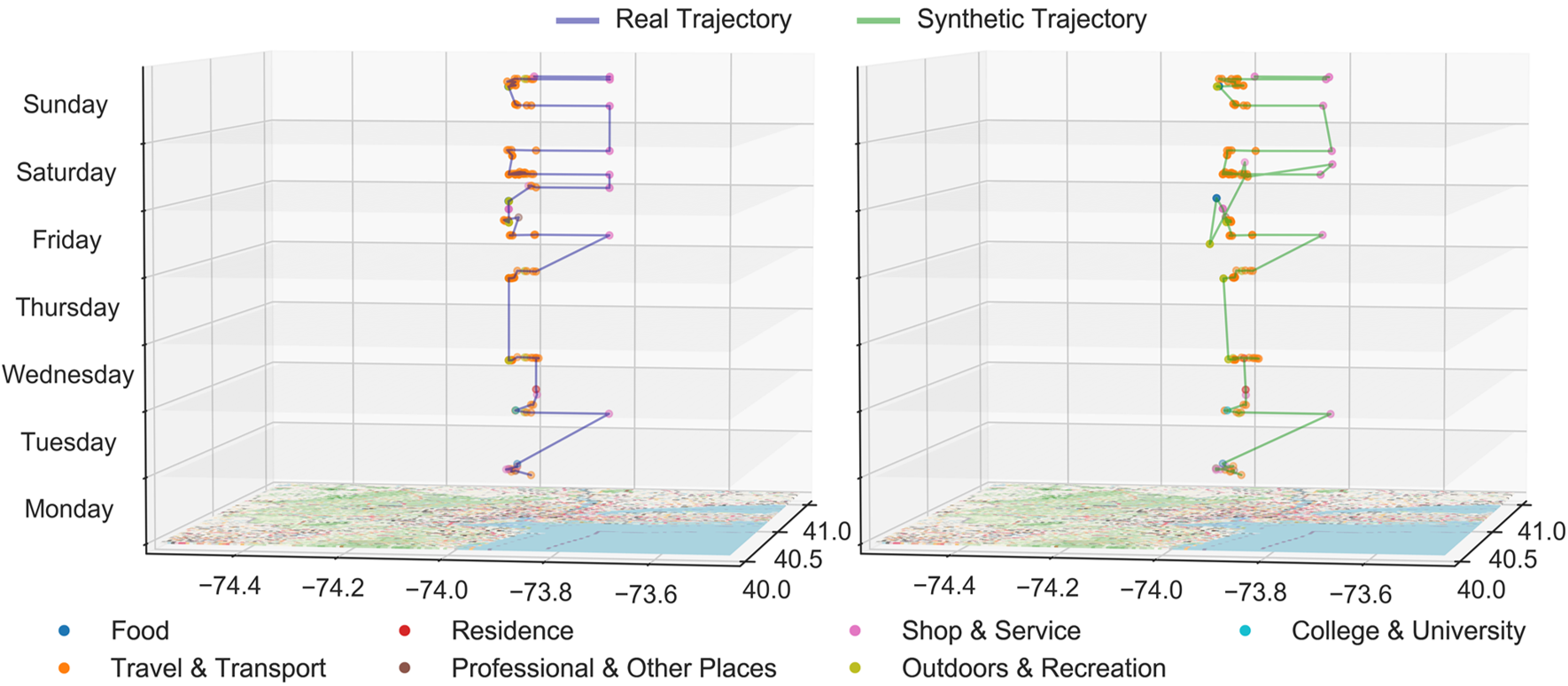}
  \caption{The visualization example of a real trajectory from the test data and its corresponding synthetic trajectory generated by our model.}
  \label{fig:trajectory_example}
\end{figure}

\paragraph*{Training and Evaluation}
We train the LSTM-TrajGAN model on the training set for 2,000 epochs with several default training hyperparameters (e.g., we use an adam optimizer with a learning rate of 0.001 and set the batch size to 256). After the training process, the trajectory data from the test set as well as random noise are then used as the input of the generator to get synthetic trajectory data. A visualization example of a real trajectory from the test data and its corresponding synthetic trajectory generated by our model is shown in Figure \ref{fig:trajectory_example} as a comparison. Next, we use the MARC (Multiple-Aspect tRajectory Classifier \cite{may2020marc}), a start-of-the-art TUL algorithm, to perform the TUL task on both the test data and our synthetic data. Same as \cite{may2020marc}, we evaluate the TUL accuracy with five commonly used metrics: ACC@1 (Top-1 Accuracy, showing the model's ability to have the correct label to be the most probable label candidate), ACC@5 (Top-5 Accuracy, showing the model's ability to have the correct label among the top 5 most probable label candidates), Macro-P (Macro Precision, the mean precision among all classes), Macro-R (Macro Recall, the mean recall among all classes), and Macro-F1 (the harmonic mean of Macro-P and Macro-R). For comparison, we also evaluate the privacy protection effectiveness of Random Perturbation (spatial filter: within 1 km; temporal filter: within 24 hours) and Gaussian Geomasking (spatial filter: standard deviation = 0.001; temporal filter: within 24 hours).

The results are shown in Table \ref{table:tulmetrics}. The higher the TUL accuracy, the worse the capability for trajectory privacy protection. One can conclude that the synthetic data generated by the LSTM-TrajGAN successfully suppress the scores in the four metrics (ACC@1, Macro-P, Macro-R, and Macro-F) from over 0.900 to around 0.400. The Top-5 Accuracy is decreased from over 0.976 to 0.722. The results show that our model can effectively prevent users from being identified by analyzing the trajectories. Additionally, Random Perturbation has limited effectiveness in protecting trajectory privacy regarding the TUL task, and Gaussian Geomasking works better while still has higher scores than our model. The results also indicate that leveraging both spatial and temporal dimensions of the trajectories simultaneously leads to better privacy-preserving performance than using only the spatial dimension.

\begin{table}[!htb]
	\centering
	\caption{The summary of the Foursquare NYC weekly trajectory dataset.}
\renewcommand{\arraystretch}{0.5}
\begin{tabular}{@{}p{3cm}p{3cm}p{4cm}@{}}
 \toprule
 \hfil\textbf{Attribute} & \hfil\textbf{Type} & \hfil\textbf{Number / Range}\\\midrule
 \hfil Trajectory ID  &\hfil integer  &\hfil 3,079\\\midrule
 \hfil User ID  &\hfil integer  &\hfil 193\\\midrule
 \hfil Latitude  &\hfil float  &\hfil (40.550852, 40.988332)\\\midrule
 \hfil Longitude  &\hfil float  &\hfil (-74.269644, -73.685767)\\\midrule
 \hfil Hour  &\hfil integer  &\hfil 24\\\midrule
 \hfil Day  &\hfil string  &\hfil 7\\\midrule
 \hfil Category  &\hfil string  &\hfil 10\\\bottomrule
\end{tabular}
\renewcommand{\arraystretch}{1}
\label{table:tuldataset}
\end{table}
\begin{table}[!htb]
	\centering
	\caption{The privacy protection effectiveness of different privacy protection methods on the TUL task (RP stands for Random Perturbation; Gaussian stands for Gaussian Geomasking).}
\renewcommand{\arraystretch}{0.5}
\begin{tabular}{@{}cccccc@{}}
 \toprule
 \textbf{Method} & \textbf{ACC@1} & \textbf{ACC@5} & \textbf{Macro-F1} & \textbf{Macro-P} & \textbf{Macro-R}\\\midrule
 Original  &0.938  &0.976  &0.925  &0.937  &0.927\\\midrule
 RP (Spatial Only)  &0.777  &0.934  &0.758  &0.806  &0.764\\\midrule
 RP (Spatial-Temporal)  &0.668  &0.888  &0.640  &0.711  &0.654\\\midrule
 Gaussian (Spatial Only)  &0.561  &0.832  &0.522  &0.573  &0.537\\\midrule
 Gaussian (Spatial-Temporal)  &0.486  &0.766  &0.431  &0.488  &0.470\\\midrule
 LSTM-TrajGAN   &\textbf{0.459}  &\textbf{0.722}  &\textbf{0.381}  &\textbf{0.429}  &\textbf{0.428}\\\midrule
\end{tabular}
\renewcommand{\arraystretch}{1}
	\label{table:tulmetrics}
\end{table}


\subsection{  Synthetic Trajectory Characteristics Analysis}
Here, we analyze the spatial and temporal characteristics and other properties of the synthetic trajectories generated by the LSTM-TrajGAN to evaluate its utility (RQ2).

\paragraph*{Spatial Characteristics}
The spatial characteristics are explored based on two metrics: the Hausdorff Distance and the Jaccard Index. The Hausdorff Distance is a metric for measuring the distance between two point sets in a metric space and has been widely used for measuring the spatial dissimilarity between two trajectories. The Jaccard Index, also known as the Intersection over Union, is an efficient metric for measuring how much the two sample sets or regions overlap, and we use this to indicate the similarity of the activity spaces between two trajectories \cite{liu2018trajgans}. We calculate the Hausdorff Distance between each pair of the original and the synthetic trajectories. Likewise, we also calculate the Jaccard Index between the convex hulls of them since the convex hull can generally represent the activity space of LBS users \cite{lee2016activity}. Table \ref{table:spatialc} presents the summary of these metrics. 

It shows that Random Perturbation has the smallest average Hausdorff Distance (0.004) and the largest average Jaccard Index (0.763), which makes sense since it only makes a limited influence on the spatial dimension of the trajectories. While such a method could preserve spatial similarity well, it sacrifices the location privacy. Our model performs better than Gaussian Geomasking on these two metrics and also better suppresses the abovementioned TUL metrics, which strikes a better balance between spatial similarity and location privacy.

\begin{table}[!htb]
	\centering
	\caption{Spatial characteristics evaluation based on Hausdorff Distance and Jaccard Index (RP stands for Random Perturbation; Gaussian stands for Gaussian Geomasking).}
\renewcommand{\arraystretch}{0.5}
\begin{tabular}{@{}ccccccccc@{}}
 \toprule
& \multicolumn{4}{c}{\textbf{Hausdorff Distance}} & \multicolumn{4}{c}{\textbf{Jaccard Index}} \\\cmidrule{2-9}
 \multirow{2}{*}[10pt]{\textbf{Method}} & \textbf{Min} & \textbf{Max} & \textbf{Std} & \textbf{Mean} & \textbf{Min} & \textbf{Max} & \textbf{Std} & \textbf{Mean}\\\midrule
 RP &0.001 &0.006  &0.001   &0.004    &0.000  &0.977  &0.194  &0.763\\\midrule
 Gaussian  &0.001  &0.034  &0.005  &0.014  &0.000 &0.933 &0.231   &0.478\\\midrule
 LSTM-TrajGAN  &0.001  &0.046  &0.006  &0.012 &0.000    &0.951  &0.234   &0.582\\\midrule
\end{tabular}
\renewcommand{\arraystretch}{1}
	\label{table:spatialc}
\end{table}

\begin{figure}[!htb]
  \centering
  \includegraphics[width=0.8\linewidth]{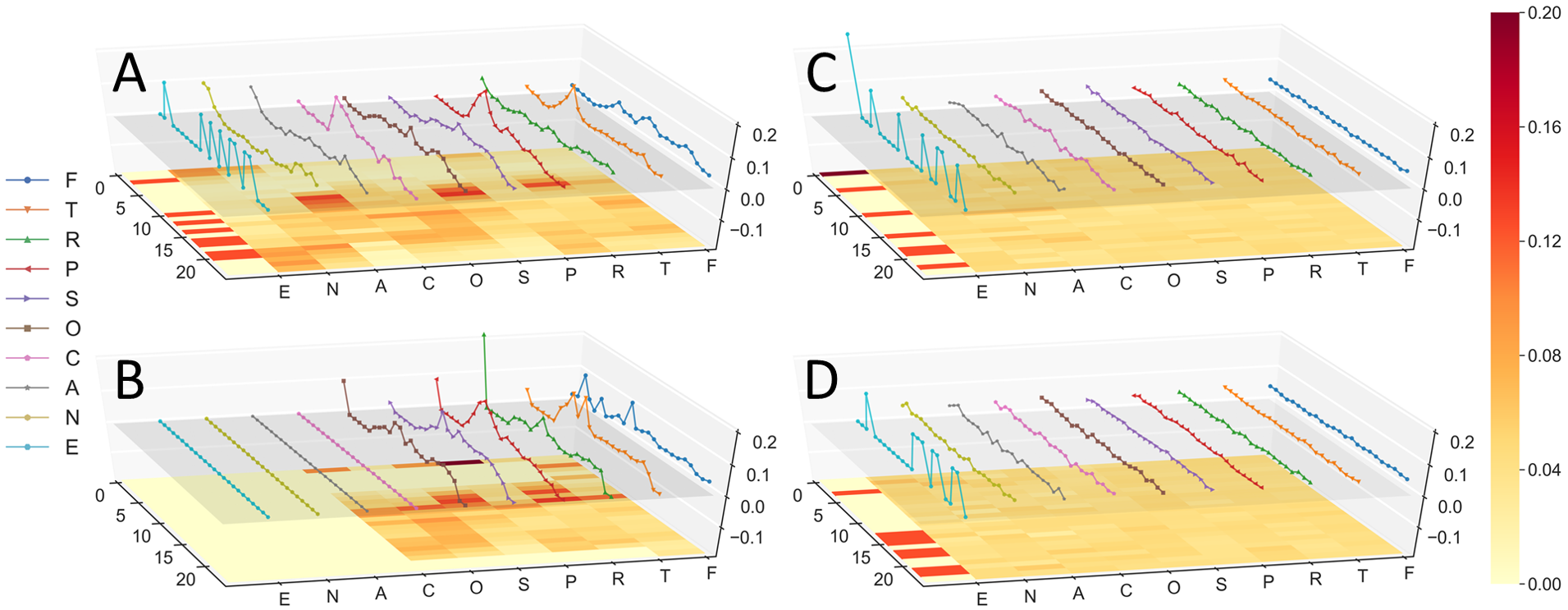}
  \caption{The hourly temporal visit probability distribution for each POI category by (A) Original data, (B) LSTM-TrajGAN, (C) Random Perturbation (within 24 hours), and (D) Gaussian Geomasking (within 24 hours) data (F: Food; T: Travel \& Transport; R: Residence; P: Professional \& Other Places; S: Shop \& Service; O: Outdoors \& Recreation; C: College \& University; A: Arts \& Entertainment; N: Nightlife Spot; E: Event).}
  \label{fig:time_probability}
\end{figure}

\paragraph*{Temporal Characteristics}
We also explore the temporal characteristics based on the visualization of two summary indicators: temporal visit probability distribution for each POI category, and overall temporal visit frequency distribution. We count the frequencies of visits to each POI category at each hour in original trajectories and the synthetic trajectories using three different approaches, and convert them into probability distribution matrices (Figure \ref{fig:time_probability}), in which the temporal patterns and the temporal similarity can be analyzed and compared.

\begin{figure}[!htb]
\captionsetup[subfigure]{justification=centering}
\centering
\begin{subfigure}{0.4\textwidth}
  \includegraphics[width=\textwidth]{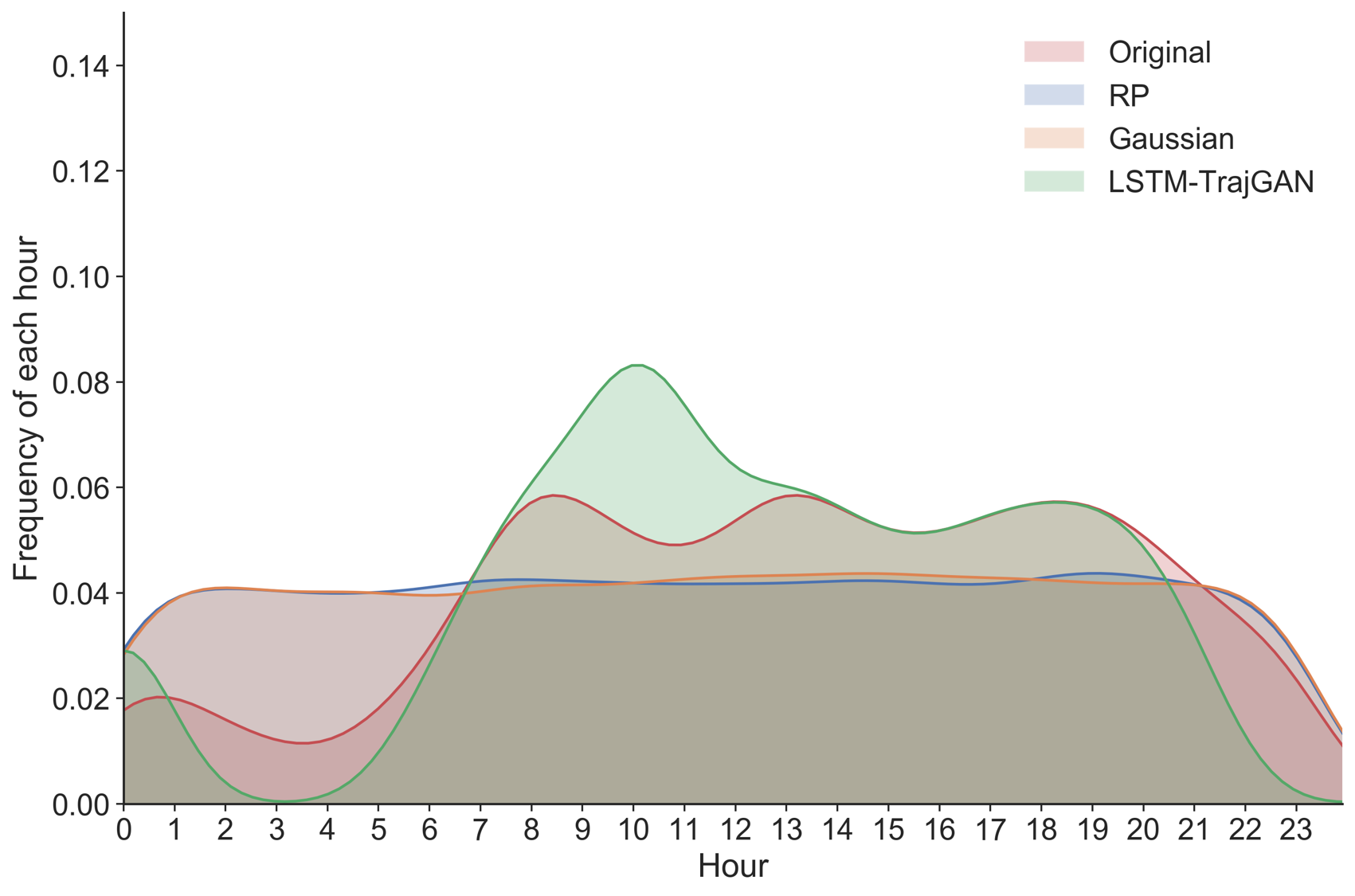}
  \caption{Overall temporal visit frequency distribution}
  \label{table:time_category_distribution:a}
\end{subfigure}
\begin{subfigure}{0.4\textwidth}
  \includegraphics[width=\textwidth]{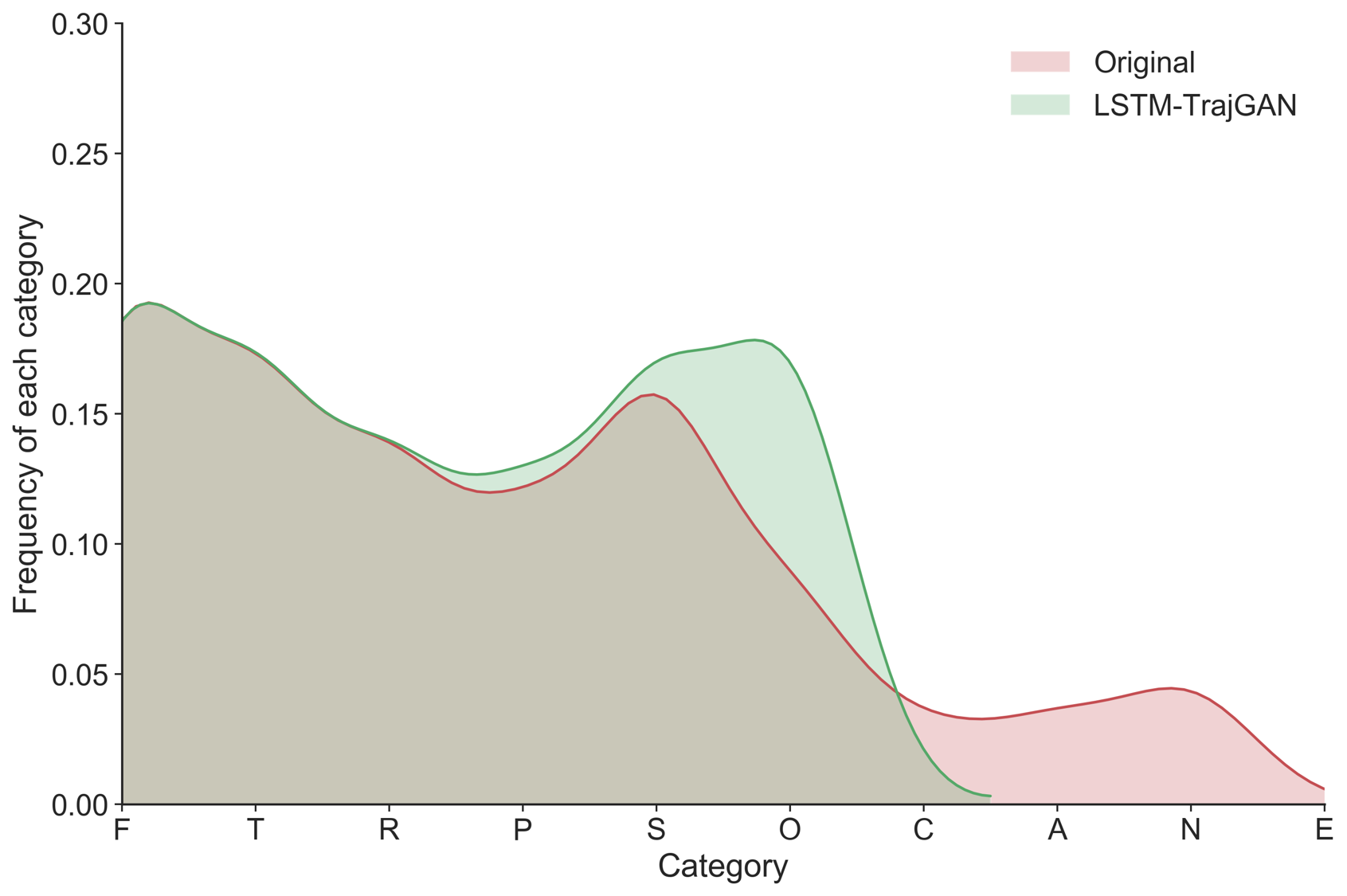}
  \caption{Overall categorical visit frequency distribution}
  \label{table:time_category_distribution:b}
\end{subfigure}
\label{table:time_category_distribution}
\caption{Overall temporal visit frequency distribution and overall categorical visit frequency distribution (RP stands for Random Perturbation; Gaussian stands for Gaussian Geomasking).}
\end{figure}

It shows that the temporal visit probability distribution from LSTM-TrajGAN shares a large commonality with that from the original data, which embodies a significant temporal similarity. Some parts of the result by LSTM-TrajGAN (i.e., categories C and E) have near zero visit probability since these categories rarely appear in training data and thus the model doesn't learn sufficient information to make intelligent predictions on them. As the comparisons, the temporal visit probability from Random Perturbation and Gaussian Geomasking show neither temporal similarity with the original data nor significant temporal patterns over 24 hours (except for the Event category).

Besides, we investigate the overall temporal and categorical visit frequency distribution (Figure \ref{table:time_category_distribution:a} and Figure \ref{table:time_category_distribution:b}). The overall temporal visit frequency distribution from our model can better fits the original data (Pearson Coefficient: 0.761) than Random Perturbation (0.536) and Gaussian Geomasking (0.535). The overall categorical visit frequency distribution also fits well (0.889). Hence, we conclude that our model generally well preserves both temporal and categorical characteristics.
\section{  Discussion}
This section discusses the factors that may affect the privacy protection effectiveness of the LSTM-TrajGAN model, and the trade-off between the privacy protection effectiveness and the utility. Finally, we discuss the limitations of our approach.

\subsection{  Factors Affecting Privacy Protection Effectiveness}
\paragraph*{Training and Optimization Settings}
We first explore how the different learning rates, loss metric functions, and random noise data affect the metric scores in the TUL task compared with the baseline setting (i.e., the learning rate = 0.001; the spatial dimension = 64; and the TrajLoss metric function during training). As shown in Table \ref{table:discusstulmetrics}, different random noise data have small influences on the metrics, which in fact contributes to the potential generalizability of the proposed approach for generating privacy-preserving trajectory data. We also found that the selection of the learning rate may have a great influence on the metrics. A higher learning rate (0.002) makes the model converge faster, generating the synthetic trajectories that have less uncertainty and share more characteristics with the original trajectories, leading to higher TUL metric scores and vice versa. Although this is not always the case, the learning rate should be carefully set to balance the trajectory utility and privacy protection effectiveness.

\begin{table}[H]
	\centering
	\caption{The metrics in the TUL task based on the synthetic trajectories by LSTM-TrajGAN using different training and optimization settings as well as different spatial embedding dimensions.}
\renewcommand{\arraystretch}{0.5}
\begin{tabular}{@{}cccccc@{}}
 \toprule
 \textbf{LSTM-TrajGAN} & \textbf{ACC@1} & \textbf{ACC@5} & \textbf{Macro-F1} & \textbf{Macro-P} & \textbf{Macro-R}\\\midrule
 Baseline  &0.459  &0.722  &0.381  &0.429  &0.428\\\midrule
 Different Random Noise  &0.466  &0.741  &0.398  &0.451  &0.436\\\midrule
 Higher Learning Rate (0.002)  &0.841  &0.959  &0.824  &0.855  &0.828\\\midrule
 Lower Learning Rate (0.00002)  &0.055  &0.157  &0.029  &0.047  &0.054\\\midrule
 Higher Spatial Dimensions (128)  &0.510  &0.811  &0.504  &0.513  &0.513\\\midrule
 Lower Spatial Dimensions (32)  &0.426  &0.703  &0.396  &0.402  &0.392\\\midrule
 TrajLoss without Spatial Loss  &0.047  &0.176  &0.030  &0.037  &0.042\\\midrule
 TrajLoss without Temporal Loss  &0.093  &0.252  &0.076  &0.119  &0.089\\\midrule
 TrajLoss without Categorical Loss  &0.354  &0.623  &0.311  &0.386  &0.346\\\midrule
 No TrajLoss  &0.010  &0.032  &0.002  &0.001  &0.007\\\midrule
\end{tabular}
\renewcommand{\arraystretch}{1}
	\label{table:discusstulmetrics}
\end{table}

In addition, we also investigate how the TrajLoss metric function contributes to the training. When removing the Spatial Loss or the Temporal Loss from the TrajLoss function, the metric scores fall dramatically, implying that the synthetic trajectories fail to preserve the spatial or temporal characteristics of the original trajectories. By comparison, removing the Categorical Loss only has a limited impact on the metric scores. Not surprisingly, removing the whole TrajLoss function results in losing spatiotemporal characteristics and thus getting the lowest TUL metric scores. We conclude that the spatial and the temporal dimensions represent the essential characteristics of a trajectory and hence need to be taken into consideration explicitly in the privacy protection approaches.

\paragraph*{Spatial Embedding}

Since the embedding of temporal attributes and categorical attributes is based on their vocabulary sizes, we mainly discuss the spatial embedding. The commonly used methods for spatial embedding are Multilayer Perceptron (MLP) and the Geohash algorithm. For example, Gupta et al. \cite{gupta2018social} use a MLP to embed the location of each person to obtain a fixed-length vector and use the vector as the input for an LSTM model to generate human trajectory. Petry et al. \cite{may2020marc} introduce a binary Geohash algorithm, in which they first use the Geohash algorithm to divide the area into grid cells and then encode the latitude and longitude as a character string, and finally convert the string into a binary fixed-length vector as the representation for the spatial dimension of each trajectory point. 

We use MLPs in the generator and the discriminator to embed the spatial dimension, but we implement them in a different way. Instead of directly embedding the coordinates, we first derive the deviations of latitudes and longitudes from the centroid of all trajectory locations, and then we embed these deviations into 64-dimensional vectors using MLP. There are two considerations: (1) On the one hand, unlike the trajectory classification task in \cite{may2020marc}, our goal is to generate synthetic trajectories, which means we need to decode the coordinates out from the hidden features in the model, and therefore using binary Geohash may lead to difficulties in learning the valid representation of coordinates, in designing the proper spatial loss, and in back-propagating the errors; and (2) On the other hand, unlike the restricted prediction area described by a Cartesian coordinate system in \cite{gupta2018social}, the prediction area in our task is on the city scale, and the difference between two GPS coordinates only appears after the decimal point. It would be a grand challenge for the model to learn and predict the coordinates with only subtle changes. As such, we standardized the coordinates to make the difference between two locations more significant for the model to learn. Recent studies also indicate that scattering the locations based on deviations may help preserve privacy \cite{gao2019exploring}.

We also explore how the spatial embedding dimensions affect the metrics in the TUL task. As is presented in Table \ref{table:discusstulmetrics}, embedding the location information into a vector with higher dimensions (e.g., 128) improves the TUL metric scores and vice versa. This makes sense since vectors in a higher-dimensional space are usually able to extract and embed more information than that in a lower-dimensional space. However, this also involves a trade-off between location accuracy and computational effort due to the limitation of physical devices.

\subsection{  The Trade-off between Privacy Protection Effectiveness and Utility}

\begin{figure}[h]
\captionsetup[subfigure]{justification=centering}
\centering
\begin{subfigure}{0.35\textwidth}
  \includegraphics[width=\linewidth]{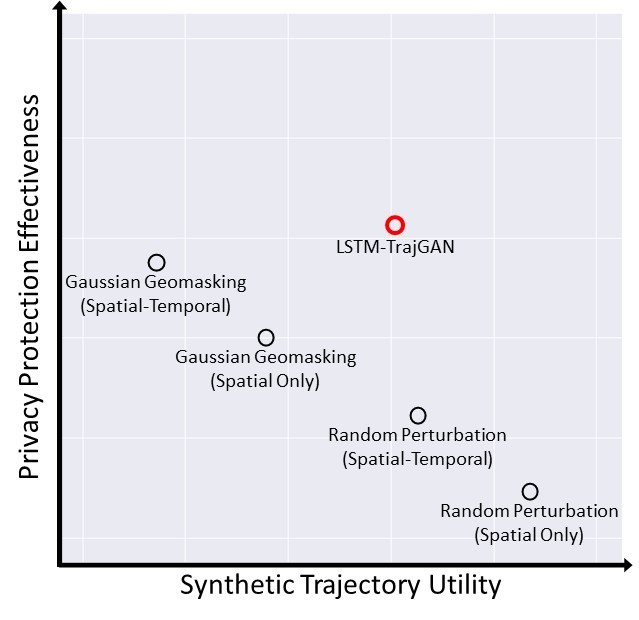}
  \caption{}
  \label{fig:tradeoff:a}
\end{subfigure}
\begin{subfigure}{0.55\textwidth}
  \includegraphics[width=\textwidth]{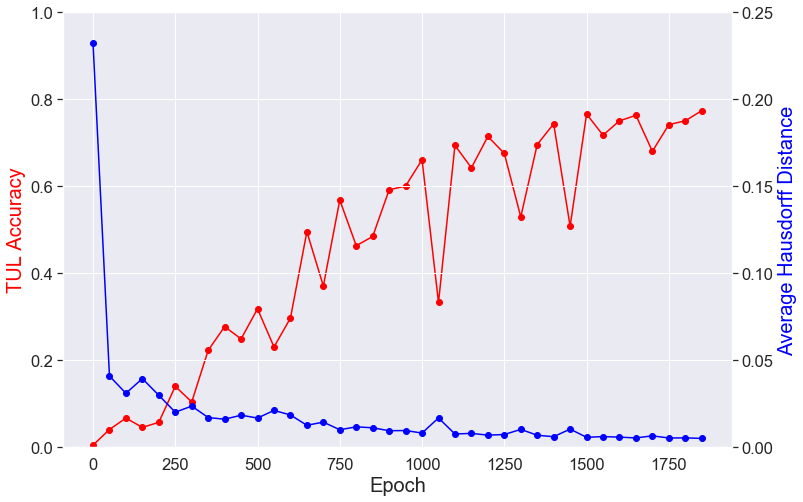}
  \caption{}
  \label{fig:tradeoff:b}
\end{subfigure}
\label{fig:tradeoff}
\caption{(a) The performance of each method in privacy protection effectiveness and utility; (b) The trade-off between the effectiveness of privacy protection (presented by TUL Top-5 Accuracy) and the preservation of spatial characteristics preservation (presented by Average Hausdorff Distance)}
\end{figure}

Generally speaking, specific trajectory analysis tasks may rely on different types of trajectory data (e.g., POI-based or road network-based) or different requirements (e.g., road extraction requires the location of each trajectory point to be precise), making it challenging to design a generic privacy protection method. However, we can evaluate a method by some specific criteria to determine its application scenarios, and even design a method based on this consideration to cover as many scenarios as possible. Inspired by the evaluation framework that involves the privacy, analytics, and uncertainty \cite{gao2019exploring}, we investigate the relationship between privacy protection effectiveness and utility. Figure \ref{fig:tradeoff:a} demonstrates the performance of each method. It is worth noting that the placement of each method is estimated from our experiment. We believe that the consideration of this relationship would help choose and design proper trajectory privacy protection methods for specific scenarios.

Sometimes the relationship between the privacy protection effectiveness and the utility is somewhat contradictory: we hope that the synthetic data are less similar to the original data to protect privacy while still preserve some similarities as good alternatives for spatiotemporal modeling or analyses. This may result in a ``catch-22 situation''. As an end-to-end deep learning model, the LSTM-TrajGAN is able to monitor and quantify this relationship during training and help to find the best-balanced parameter settings. For example, as training progresses, the TUL accuracy (Top-5 Accuracy) increases while the Average Hausdorff Distance decreases (Figure \ref{fig:tradeoff:b}). Carefully selecting the model weight from different epochs based on this relationship could ensure that the synthetic trajectories preserve spatiotemporal characteristics to some extent while maintaining a low TUL accuracy as needed, thereby balancing the privacy protection effectiveness and the utility of synthetic trajectories.

\subsection{  Limitations}
Several limitations exist in our current approach. First, compared to traditional geomasking techniques that blur the existing trajectories, our deep learning model that generates new trajectories leads to a much higher computational effort and also needs an additional training process before its deployment in applications. Second, we focus on the TUL task and analyzed spatial and temporal characteristics of the synthetic trajectories, which reflects their potential for privacy-preserving trajectory analysis, but more specific evaluations are not investigated yet. Third, our model generates only the synthetic trajectories that have the same length as the original trajectories. Finally, our model currently focuses on city-scale trajectories, and the deviation-based location representation may not be suitable for global-scale trajectories. These limitations will be further explored in our future work.
\section{  Conclusion and Future Work}

This research proposes a novel LSTM-TrajGAN approach, i.e., a deep learning model that combines the LSTM recurrent neural network and the GAN structure to generate privacy-preserving synthetic trajectories for trajectory data publication. We utilize the idea of adversarial training in the model design, train our model on a Foursquare NYC weekly trajectory dataset, and evaluate its privacy protection effectiveness in the TUL task. To answer the two research questions we posed at the beginning of this research, the results show that (RQ1) our model can generate the spatial-temporal synthetic trajectories that prevent the trajectory creators (i.e., users) from being re-identified to certain degree and (RQ2) keep some spatial, temporal, and thematic characteristics of the original trajectories. Additionally, the results show that the model has the potentials for supporting further spatial or temporal analyses. Lastly, we explored the factors affecting the privacy protection effectiveness and discussed the trade-off between model effectiveness and utility in general. The design of a new loss function TrajLoss offers new insights into the development of spatially explicit artificial intelligence techniques for advancing GeoAI \cite{janowicz2019geoai}.

Our future work will focus on improving the trajectory similarity loss metric function, extending our framework to global-scale trajectory datasets, generating custom variable-length synthetic trajectory data, exploring potential privacy attack and defense strategies, and evaluating the privacy protection effectiveness and utility of our model in other trajectory data mining and analysis tasks.


\bibliography{geoprivacy}

\end{document}